%% file: example_paper.tex
\documentclass[runningheads]{llncs}

\usepackage{microtype}
\usepackage{graphicx}
\usepackage{booktabs} % for professional tables
\usepackage{epsfig}
\usepackage{graphicx}
\usepackage{lipsum}
\usepackage{subcaption}
\usepackage{graphicx}
\usepackage{amsmath,amssymb} % define this before the line numbering.
\usepackage{color}
\usepackage{adjustbox}
\usepackage{multirow}
\usepackage{wrapfig}
\usepackage{booktabs}
\usepackage{comment}

\usepackage{hyperref}
\usepackage{mathtools, nccmath}
\input{math_commands.tex}

\begin{document}

\DeclarePairedDelimiter{\nint}\lfloor\rceil

\pagestyle{headings}
\mainmatter

\input listofitems
\let\cpar\relax
\def\Centerline#1{%
  \setsepchar{\cpar}%
  \readlist\clarg{#1}%
  \foreachitem\z\in\clarg[]{\centerline{\z}}%
}

\title{One Weight Bitwidth to Rule Them All} % Replace with your title

\begin{comment}
\titlerunning{ECCV-20 submission ID \ECCVSubNumber} 
\authorrunning{ECCV-20 submission ID \ECCVSubNumber} 
\author{Anonymous ECCV submission}
\institute{Paper ID \ECCVSubNumber}
\end{comment}
\newcommand\rudy[1]{{\color{black}{#1}}}
\newcommand\dm[1]{{\color{blue}{#1}}}

% \begin{comment}
\titlerunning{One Weight Bitwidth to Rule Them All}
\author{Ting-Wu Chin\inst{1} \and
Pierce I-Jen Chuang\inst{2} \and
Vikas Chandra\inst{2} \and
Diana Marculescu\inst{1,3}}
\authorrunning{Chin et al.}
\institute{Eletrical and Computer Engineering, Carnegie Mellon University\\
\email{\{tingwuc,dianam\}@cmu.edu}\and
Facebook Inc.\\
\email{\{pichuang,vchandra\}@fb.com}\and
Eletrical and Computer Engineering, The University of Texas at Austin\\
\email{\{dianam\}@utexas.edu}}
% \end{comment}
%******************
\maketitle

\begin{abstract}
Weight quantization for deep ConvNets has shown promising results for applications such as image classification and semantic segmentation and is especially important for applications where memory storage is limited. However, when aiming for quantization without accuracy degradation, different tasks may end up with different bitwidths. This creates complexity for software and hardware support and the complexity accumulates when one considers mixed-precision quantization, in which case each layer's weights use a different bitwidth. Our key insight is that optimizing for the least bitwidth subject to no accuracy degradation is not necessarily an optimal strategy. This is because one cannot decide optimality between two bitwidths if one has smaller model size while the other has better accuracy. In this work, we take the first step to understand if some weight bitwidth is better than others by aligning all to the same model size using a width-multiplier. Under this setting, somewhat surprisingly, we show that using a single bitwidth for the whole network can achieve better accuracy compared to mixed-precision quantization targeting zero accuracy degradation when both have the same model size. In particular, our results suggest that when the number of channels becomes a target hyperparameter, a single weight bitwidth throughout the network shows superior results for model compression.

\end{abstract}
\section{Introduction}\label{sec:intro}
Recent success of ConvNets in computer vision applications such as image classification and semantic segmentation has fueled many important applications in storage-constrained devices, \emph{e.g.}, virtual reality headsets, drones, and IoT devices. As a result, improving the parameter-efficiency (the top-1 accuracy to the parameter counts ratio) of ConvNets while maintaining their attractive features (\emph{e.g.}, accuracy for a task) has gained tremendous research momentum recently.

Among the efforts of improving ConvNets' efficiency, weight quantization was shown to be an effective technique~\cite{zhou2016dorefa,zhou2017incremental,hou2018lossaware,Ding_2019_CVPR}. The majority of research efforts in quantization has targeted quantization algorithms for finding the lowest possible weight bitwidth without compromising the figure-of-merit (\emph{i.e.}, accuracy). Mixed-precision quantization methods, which allow different bitwidths to be selected for different layers in the network, have recently been proposed to further compress deep ConvNets~\cite{wang2019haq,wu2018mixed,dong2019hawq}. Nevertheless, having different bitwidths for different layers greatly increases the neural network implementation complexity from both hardware and software perspectives. For example, hardware and software implementations optimized for executing an 8~bits convolution are sub-optimal for executing a 4~bits convolution, and vice versa.

To minimize the efforts of hardware and software support, it is natural to wonder: ``Is some weight bitwidth better than others?'' However, this is an ill-posed problem as one cannot decide optimality between two bitwidths if one has smaller model size while the other has better accuracy. This work takes a first step towards understanding if some bitwidth is better than other bitwidths under a given model size constraint. Given the multi-objective nature of the problem, we need to align different bitwidths to the same model size to further decide the optimality for the bitwidth selection. To realize model size alignment for different bitwidths, we use the width-multiplier\footnote{Width-multiplier grows or shrinks the number of channels across the layers with identical proportion for a certain network, \emph{e.g.}, grow the number of channels for all the layers by $2\times$.}~\cite{howard2017mobilenets} as a tool to compare the performance of different weight bitwidths under \emph{the same model size}.

With this setting, we find that there exists some weight bitwidth that consistently outperforms others across different model sizes when both are considered under a given model size constraint. This suggests that one can decide the optimal bitwidth for small model sizes to save computing cost and the result generalizes to large model sizes\footnote{Note that we use width-multiplier to scale model across different sizes.}. Additionally, we show that the optimal bitwidth of a convolutional layer negatively correlates to the convolutional kernel fan-in. As an example, depth-wise convolutional layers turn to have optimal bitwidth values that are higher than that of all-to-all convolutions. We further provide a theoretical reasoning for this phenomenon. These findings suggest that architectures such as VGG and ResNets are more parameter-efficient when they are wide and use binarized weights. On the other hand, networks such as MobileNets~\cite{howard2017mobilenets} might require different weight bitwidths for all-to-all convolutions and depth-wise convolutions. Somewhat surprisingly, we find that on ImageNet, under a given model size constraint, a single bitwidth for both ResNet-50 and MobileNetV2 can outperform mixed-precision quantization using reinforcement learning~\cite{wang2019haq} that targets minimum total bitwidth without accuracy degradation. This suggests that searching for the minimum bitwidth configuration that does not introduce accuracy degradation without considering other hyperparameters affecting model size is a sub-optimal strategy. Our results suggest that when the number of channels becomes one of the hyperparameters under consideration, a single weight bitwidth throughout the network shows great potential for model compression.

In summary, we systematically analyze the model size and accuracy trade-off considering both weight bitwidths and the number of channels for various modern networks architectures (variants of ResNet, VGG, and MobileNet) and datasets (CIFAR and ImageNet) and have the following contributions:
\begin{itemize}
    \item We empirically show that when allowing the network width to vary, lower weight bitwidths outperform higher ones in a Pareto sense (accuracy \textit{vs.} model size) for networks with standard convolutions. This suggests that for such ConvNets, further research on wide binary weight networks is likely to identify better network configurations which will require further hardware/software platform support.
    \item We empirically show that the optimal bitwidth of a convolutional layer negatively correlates to the convolutional kernel fan-in and provide theoretical reasoning for such a phenomenon. This suggests that one could potential categorize ConvNets based on the convolutional kernel fan-in when designing the corresponding bitwidth support from both software and hardware.
    \item We empirically show that one can achieve a more accurate model (under a given model size) by using a single bitwidth when compared to mixed-precision quantization that uses deep reinforcement learning to search for layer-wise weight precision values. Moreover, the results are validated on a large-scale dataset, \emph{i.e.}, ImageNet.
\end{itemize}

The remainder of the paper is organized as follows. Section~\ref{sec:related} discusses related work. Section~\ref{sec:method} discusses the methodology used to discover our findings. Section~\ref{sec:exp} discusses our experiments for all our findings. In particular, Section~\ref{sec:first_finding} shows that some bitwidth can outperform others consistently across model sizes when both are compared under the same model size constraint using width-multipliers. Section~\ref{sec:second_finding} discusses how fan-in channel count per convolutional kernel affects the resilience of quantization for convolution layers, which further affects the optimal bitwidth for a convolution layer. Section~\ref{sec:third_finding} scales up our experiments to ImageNet and demonstrates that a single weight bitwidth manages to outperform mixed-precision quantization given the same model size. Section~\ref{sec:conclusion} concludes the paper.

\section{Related Work}\label{sec:related}
Several techniques for improving the efficiency of ConvNets have been recently proposed. For instance, pruning removes the redundant connections of a trained neural network~\cite{zhuang2018discrimination,ye2018rethinking,theis2018faster,li2016pruning,frankle2018the,chin2019legr,Yu_2018_CVPR}, neural architecture search (NAS) tunes the number of channels, size of kernels, and depth of a network~\cite{tan2018mnasnet,stamoulis2019single,cai2018proxylessnas,stamoulis2018designing}, and convolution operations can be made more efficient via depth-wise convolutions~\cite{howard2017mobilenets}, group convolutions~\cite{huang2017condensenet,zhao2019building}, and shift-based convolutions~\cite{he2019addressnet,wu2018shift}. In addition to the aforementioned techniques, network quantization introduces an opportunity for hardware-software co-design to achieve better efficiency for ConvNets.

There are in general two directions for weight quantization in prior literature, post-training quantization~\cite{nagel2019data,meller19a,zhao19c,sheng2018quantization} and quantization-aware training~\cite{rastegari2016xnor,zhu2016trained,Jacob_2018_CVPR,Jung_2019_CVPR,Yuan_2019_CVPR,hou2018lossaware,choi2018bridging}. The former assumes training data is not available when quantization is applied. While being fast and training-data-free, its performance is worse compared to quantization-aware training. In contrast, our work falls under the category of quantization-aware training.

In quantization-aware training, \cite{rastegari2016xnor} introduces binary neural networks, which lead to significant efficiency gain by replacing multiplications with XNOR operations at the expense of significant accuracy degradation. Later, \cite{zhu2016trained} propose ternary quantization and \cite{zhou2016dorefa,Jacob_2018_CVPR} bridge the gap between floating-point and binarized neural networks by introducing fixed-point quantization. Building upon prior art, the vast majority of existing work focuses on reducing the accuracy degradation by improving the training strategy~\cite{zhou2017incremental,Yang_2019_CVPR,louizos2018relaxed,Ding_2019_CVPR} and developing better quantization schemes~\cite{Jung_2019_CVPR,wang2019haq,Yuan_2019_CVPR}. However, prior art has studied quantization by fixing the network architecture, which may lead to a sub-optimal bitwidth selection in terms of parameter-efficiency (the top-1 accuracy to the parameter counts ratio).

Related to our work,~\cite{mishra2018wrpn} have also considered the impact of channel count in quantization. In contrast, our work has the following novel features. First, we find that in ConvNets with standard convolutions, \emph{a lower bitwidth outperforms higher ones under a given model size constraint}. Second, we find that the Pareto optimal bitwidth negatively correlates to the convolutional kernel fan-in and we provide theoretical insights for it. \rudy{Last, we show that a single weight bitwidth can outperform \emph{mixed-precision} quantization on ImageNet for ResNet50 and MobileNetV2.}

\section{Methodology}\label{sec:method}
In this work, we are interested in comparing different bitwidths under a given model size. To do so, we make use of the width-multiplier to scale the models. To be precise in the following discussion, we define an ordering relation across bitwidths as follows:

\begin{definition}[bitwidth ordering]\label{def:bitwidth}
We say bitwidth $A$ is better than bitwidth $B$ for a network family $\mathcal{F}$, if,
$$Acc(N(A,s)) > Acc(N(B,s))~~\forall s,$$
where $Acc(\cdot)$ evaluates the validation accuracy of a network, $N(A,s)$ produces a network in $\mathcal{F}$ that has bitwidth $A$ and model size of $s$ by using width-multiplier.
\end{definition}

With Definition~\ref{def:bitwidth}, we can now compare weight bitwidths for their parameter-efficiency.

\subsection{Quantization}

\rudy{This work focuses on weight quantization and} we use a straight-through estimator~\cite{bengio2013estimating} to conduct quantization-aware training. Specifically, for bitwidth values larger than 2 bit ($b>2$), we use the following quantization function for weights during the forward pass:
\begin{equation}
    \begin{split}
        Q(\mW_{i,:}) &= \nint{\frac{clamp(\mW_{i,:}, -\eva_i, \eva_i)}{\evr_i}}\times \evr_i,~~\evr_i = \frac{\eva_i}{2^{b-1}-1}
    \end{split}
    \label{eq:quant-4bit}
\end{equation}
where
\[clamp(w,min,max) = \begin{cases}
      w, & \text{if}\ min\leq w \leq max \\
      min, & \text{if}\ w< min \\
      max & \text{if}\ w>max
    \end{cases}
\]
and $\nint{\cdot}$ denotes the round-to-nearest-neighbor function, $\mW\in \R^{C_{out}\times d},~d=C_{in} K_w K_h$ denotes the real-value weights for the $i^{\text{th}}$ output filter of a convolutional layer that has $C_{in}$ channels and $K_w \times K_h$ kernel size. $\va \in \R^{C_{out}}$ denotes the vector of clipping factors which are selected to minimize $\lVert Q(\mW_{i,:}) - \mW_{i,:} \rVert^2_2$ by assuming $\mW_{i,:}~\sim \mathcal{N}(0, \sigma^2\mI)$. More details about the determination of $\eva_i$ is in Appendix~\ref{app:alpha}. 

For special cases such as 2~bits and 1~bit, we use schemes proposed in prior literature. Specifically, let us first define:

\begin{equation}
    \bar{|\mW_{i,:}|} = \frac{1}{d}\sum_{j=1}^d |\mW_{i,j}|.
\end{equation}

For 2 bit, we follow trained ternary networks~\cite{zhu2016trained} and define the quantization function as follows:
\begin{equation}
    \begin{split}
        Q(\mW_{i,:}) &= \left(sign(\mW_{i,:}) \odot \mM_{i,j}\right)\times \left( \bar{|\mW_{i,:}|} \right) \\
        \mM_{i,j}&=\begin{cases}
            0, & \mW_{i,j} < 0.7\bar{|\mW_{i,:}|}.\\
            1, & otherwise.
          \end{cases}
    \end{split}
    \label{eq:quant-2bit}
\end{equation}

For 1 bit, we follow DoReFaNets~\cite{zhou2016dorefa} and define the quantization function as follows:
\begin{equation}
    \begin{split}
        Q(\mW_{i,:}) &= sign(\mW_{i,:}) \times \left( \bar{|\mW_{i,:}|} \right).
    \end{split}
    \label{eq:quant-1bit}
\end{equation}

For the backward pass for all the bitwidths, we use a straight-through estimator as in prior literature to make the training differentiable. That is,
\begin{equation}
    \begin{split}
        \frac{\partial Q(\mW_{i,:})}{\partial \mW_{i,:}} = \mI.
    \end{split}
\end{equation}

In the sequel, we quantize the \emph{first and last layers to 8~bits}. They are fixed throughout the experiments. We note that it is a common practice to leave the first and the last layer \emph{un-quantized}~\cite{zhou2016dorefa}, however, we find that using 8~bits can achieve comparable results to the floating-point baselines.

\rudy{As for activation, we use the technique proposed in~\cite{Jacob_2018_CVPR} and use 4~bits for CIFAR-100 and 8~bits for ImageNet experiments. The activation bitwidths are chosen such that the quantized network has comparable accuracy to the floating-point baselines.}

\subsection{Model size}
The size of the model ($C_{size}$) is defined as:
\begin{equation}
    \begin{split}
        \mathcal{C}_{size} =\sum_{i=1}^{O} b(i) C_{in}(i) K_w(i) K_h(i)
    \end{split}
\end{equation}
where $O$ denotes the total number of filters, $b(i)$ is the bitwidth for filter $i$, $C_{in}(i)$ denotes the number of channels for filter $i$, and $K_w(i)$ and $K_h(i)$ are the kernel height and width for filter $i$.

\section{Experiments}\label{sec:exp}
We conduct all our experiments on image classification datasets including CIFAR-100~\cite{krizhevsky2009learning} and ImageNet. All experiments are trained from scratch to ensure different weight bitwidths are trained equally long. While we do not start from a pre-trained model, we note that our baseline fixed-point models (\emph{i.e.}, 4~bits for CIFAR and 8~bits for ImageNet) have accuracy comparable to their floating-point counterparts. For all the experiments on CIFAR, we run the experiments three times and report the mean and standard deviation.

\subsection{Training hyper-parameters}\label{sec:hyper}
For CIFAR, we use a learning rate of 0.05, cosine learning rate decay, linear learning rate warmup (from 0 to 0.05) with 5 epochs, batch size of 128, total training epoch of 300, weight decay of $5e^{-4}$, SGD optimizer with Nesterov acceleration and 0.9 momentum.

For ImageNet, we have identical hyper-parameters as CIFAR except for the following hyper-parameters batch size of 256, 120 total epochs for MobileNetV2 and 90 for ResNets, weight decay $4e^{-5}$, and 0.1 label smoothing.

\subsection{bitwidth comparisons}\label{sec:first_finding}
In this subsection, we are primarily interested in the following question:

\smallskip
\rudy{\Centerline{\textit{\textbf{When taking network width into account, does one bitwidth}} \cpar \textit{\textbf{consistently outperform others across model sizes?}}}}
\smallskip

To our best knowledge, this is an open question and we take a first step to answer this question empirically. \rudy{If the answer is affirmative}, it may be helpful to focus the software/hardware support on the better bitwidth when it comes to parameter-efficiency. We consider three kinds of commonly adopted ConvNets, namely, ResNets with basic block~\cite{he2016deep}, VGG~\cite{simonyan2014very}, and MobileNetV2~\cite{sandler2018mobilenetv2}. These networks differ in the convolution operations, connections, and filter counts. For ResNets, we explored networks from 20 to 56 layers in six layer increments. For VGG, we investigate the case of eleven layers. Additionally, we also study MobileNetV2, which is a mobile-friendly network. We note that we modify the stride count in of the original MobileNetV2 to match the number of strides of ResNet for CIFAR. The architectures that we introduce for the controlled experiments are discussed in detail in Appendix~\ref{app:arch}.

For CIFAR-100, we only study weight bitwidths below 4 since it achieves performance comparable to its floating-point counterpart. Specifically, we consider 4~bits, 2~bits, and 1~bit weights. To compare different weight bitwidths using Definition~\ref{def:bitwidth}, we use the width-multiplier to align the model size among them. For example, one can make a 1-bit ConvNet twice as wide to match the model size of a 4-bit ConvNet~\footnote{Increase the width of a layer increases the number of output filters for that layer as well as the number of channels for the subsequent layer. Thus, number of parameters and number of operations grow approximately quadratically with the width-multiplier.}. For each of the networks we study, we sweep the width-multiplier to consider points at multiple model sizes. Specifically, for ResNets, we investigate seven depths, four model sizes for each depth, and three bitwidths, which results in $7\times 4\times 3\times 3$ experiments. For both VGG11 and MobileNetV2, we consider eight model sizes and three bitwidths, which results in $2\times 8\times 3\times 3$ experiments. 

\begin{figure}[t]
    \centering
    \begin{subfigure}[t]{0.32\textwidth}
        \centering
        \includegraphics[width=0.95\linewidth]{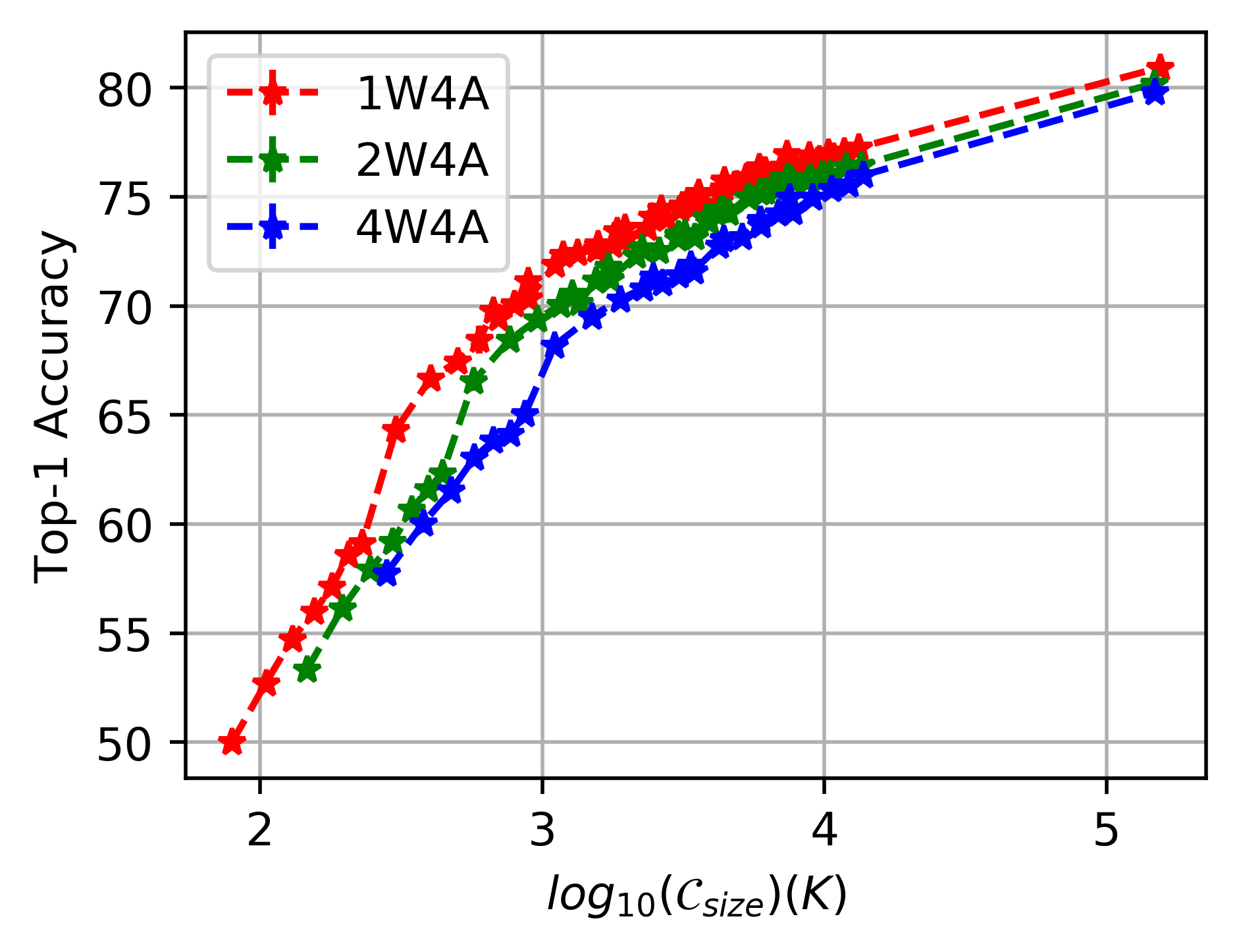}
        \caption{ResNets (20 to 56 layers in increments of 6)}\label{fig:cifar_resnet}
    \end{subfigure}%
    ~ 
    \begin{subfigure}[t]{0.32\textwidth}
        \centering
        \includegraphics[width=0.95\linewidth]{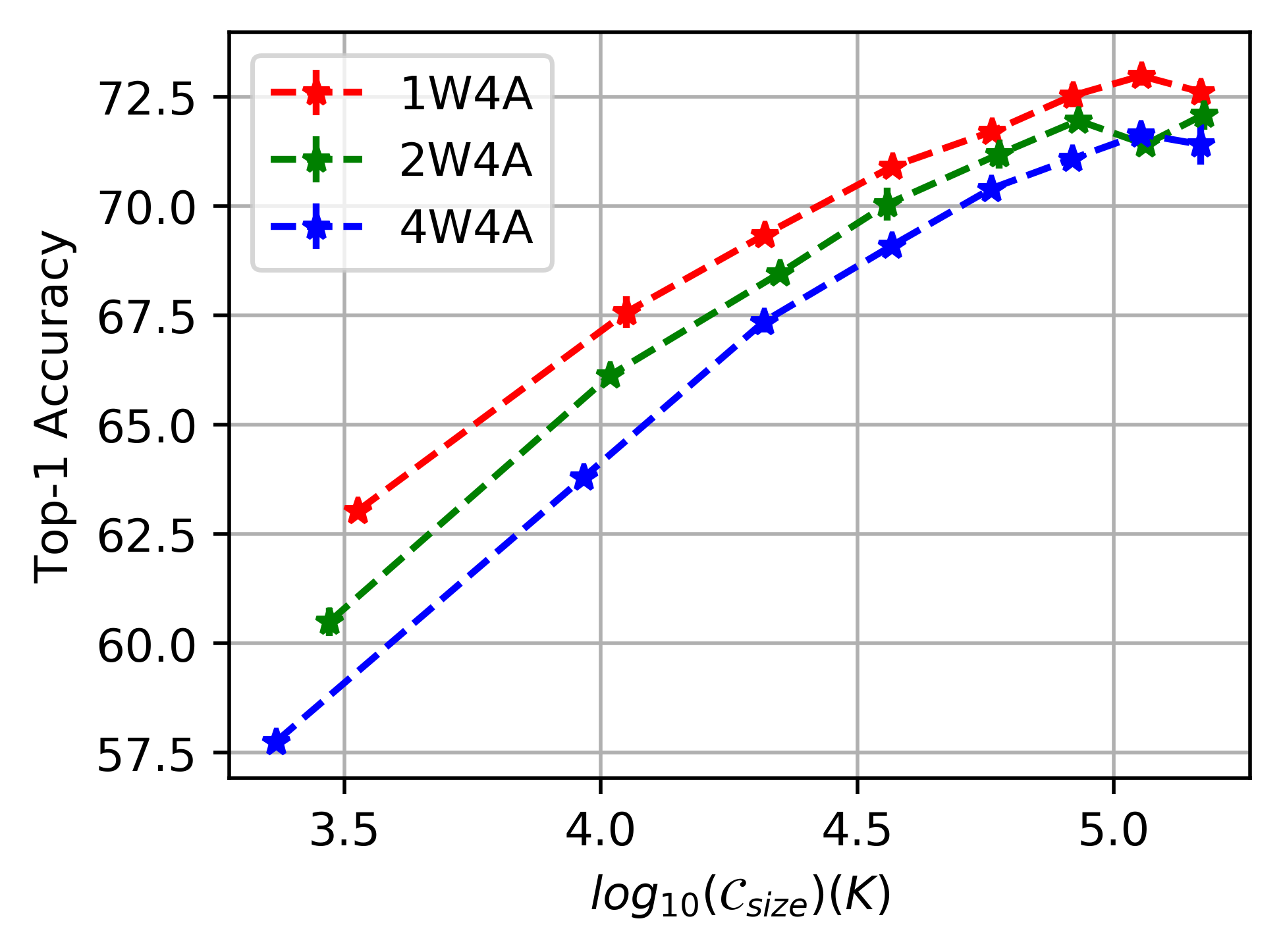}
        \caption{VGG11}\label{fig:cifar_vggnet}
    \end{subfigure}
    ~ 
    \begin{subfigure}[t]{0.32\textwidth}
        \centering
        \includegraphics[width=0.95\linewidth]{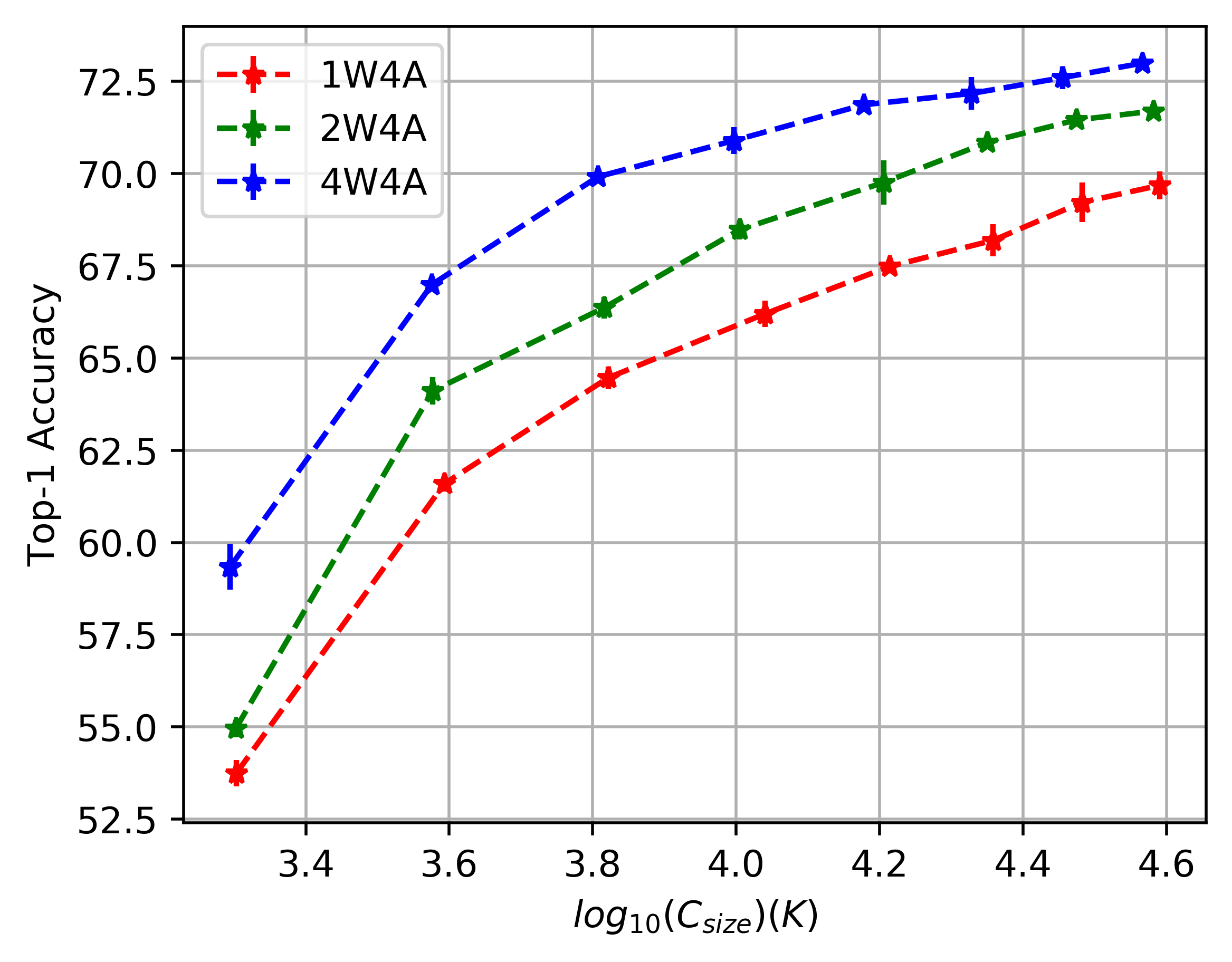}
        \caption{MobileNetV2}\label{fig:cifar_mbnet}
    \end{subfigure}
    \caption{Some bitwidth is consistently better than other bitwidths across model sizes. $\mathcal{C}_{size}$ denotes model size. $x$W$y$A denotes $x$-bit weight quantization and $y$-bit activation quantization. The experiments are done on the CIFAR-100 dataset. For each network, we sweep the width-multiplier to cover points at multiple model sizes. For each dot, we plot the mean and standard deviation of three random seeds. The standard deviation might not be visible due to little variances.}\label{fig:cifar_pareto}
\end{figure}

As shown in Fig.~\ref{fig:cifar_pareto}, across the three types of networks we study, there exists some bitwidth that is better than others. That is, the answer to the question we raised earlier in this subsection is affirmative. For ResNets and VGG, this value is 1~bit. In contrast, for MobileNetV2, it is 4~bits. The results for ResNets and VGG are particularly interesting since lower weight bitwidths are better than higher ones. In other words, binary weights in these cases can achieve the best accuracy and model size trade-off. On the other hand, MobileNetV2 exhibits a different trend where higher bitwidths are better than lower bitwidths up to 4~bits\footnote{However, not higher than 4 bits since the 4-bit model has accuracy comparable to the floating-point model.}.

\subsection{ConvNet architectures and quantization}\label{sec:second_finding}
While there exists an ordering among different bitwidths as shown in Fig.~\ref{fig:cifar_pareto}, it is not clear what determines the optimal weight bitwidth. To further uncover the relationship between ConvNet's architectural parameters and its optimal weight bitwidth, we ask the following questions.

\smallskip
\Centerline{\textit{\textbf{What architectural components determine the MobileNetV2}} \cpar \textit{\textbf{optimal weight bitwidth of 4~bits as opposed to 1~bit?}}}
\smallskip

\begin{figure}[t!]
    \centering
    \begin{subfigure}[t]{0.23\textwidth}
        \centering
        \includegraphics[width=1\linewidth]{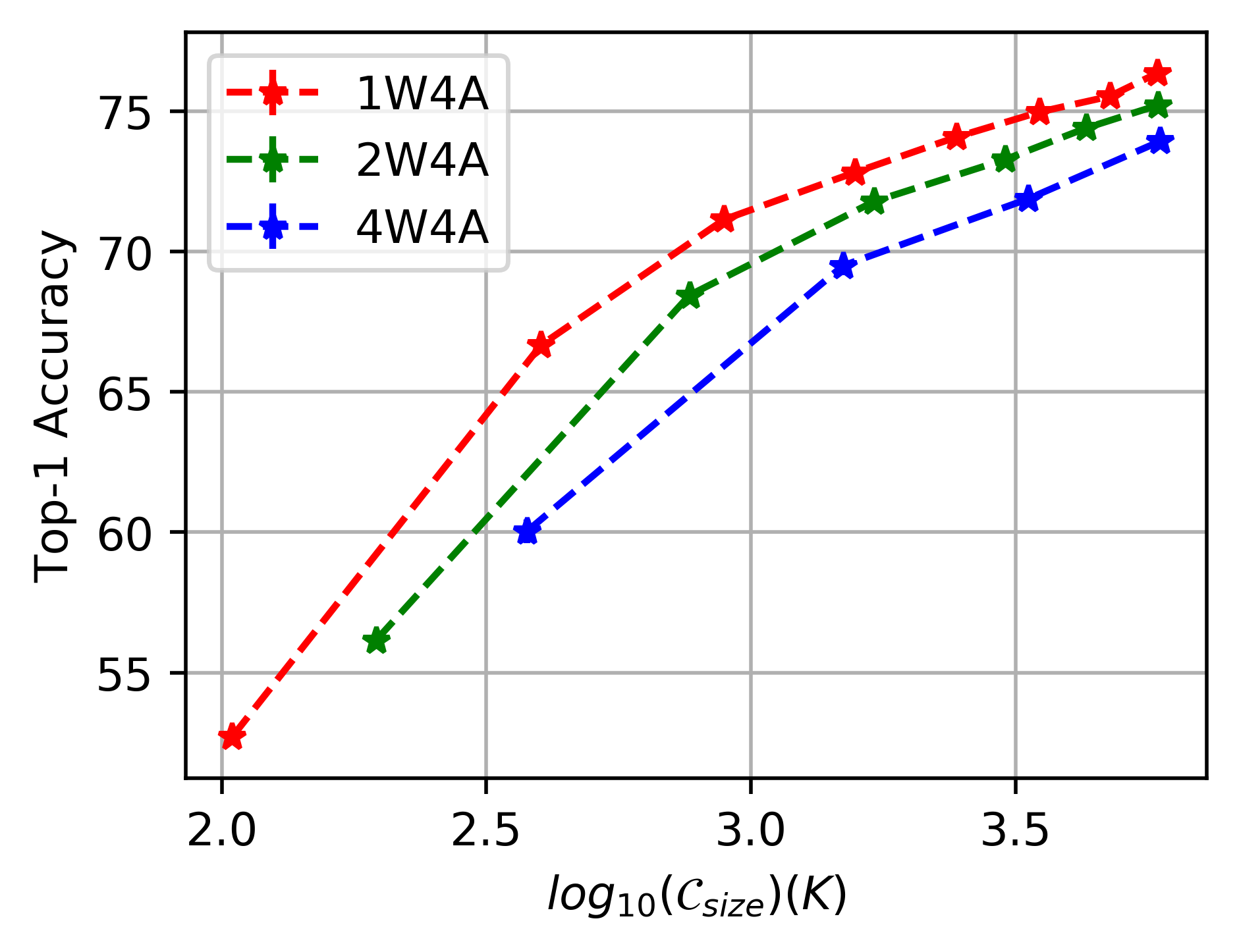}
        \caption{ResNet26}\label{fig:resnet26}
    \end{subfigure}
    ~
    \begin{subfigure}[t]{0.23\textwidth}
        \centering
        \includegraphics[width=1\linewidth]{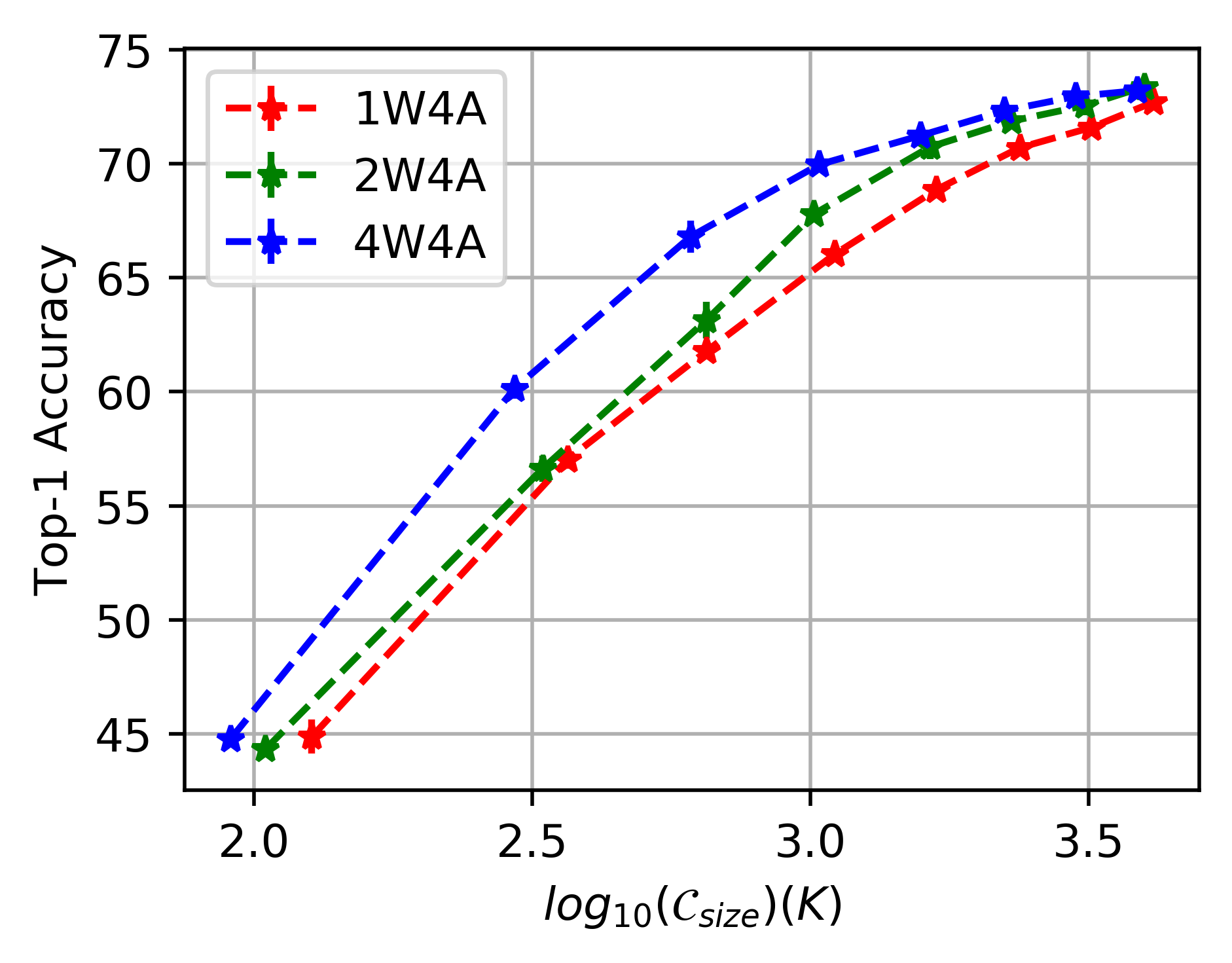}
        \caption{Inv-ResNet26}\label{fig:invresnet}
    \end{subfigure}
    ~
    \begin{subfigure}[t]{0.23\textwidth}
        \centering
        \includegraphics[width=0.65\linewidth]{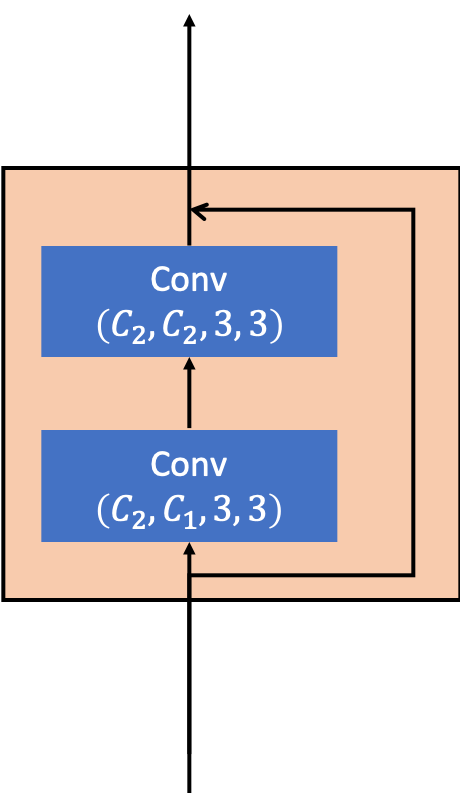}
        \caption{Basic block}\label{fig:bb}
    \end{subfigure}%
    ~ 
    \begin{subfigure}[t]{0.23\textwidth}
        \centering
        \includegraphics[width=0.65\linewidth]{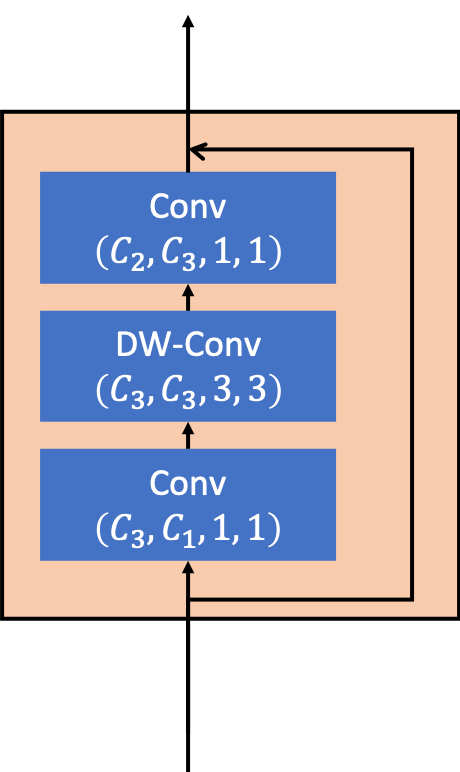}
        \caption{Inverted residual block}\label{fig:irb}
    \end{subfigure}
    \caption{The optimal bitwidth for ResNet26 changes from 1 bit (a) to 4 bit (b) when the building blocks change from basic blocks (c) to inverted residual blocks (d). $\mathcal{C}_{size}$ in (a) and (b) denotes model size. ($C_{out}$, $C_{in}$, $K$, $K$) in (c) and (d) indicate output channel count, input channel count, kernel width, and kernel height of a convolution.}
\end{figure}

As it can be observed in Fig.~\ref{fig:cifar_pareto}, MobileNetV2 is a special case where the higher bitwidth is better than lower ones. When comparing MobileNetV2 to the other two networks, there are many differences, including how convolutions are connected, how many convolutional layers are there, how many filters in each of them, and how many channels for each convolution. To narrow down which of these aspects result in the reversed trend compared to the trend exhibits in ResNets and VGG, we first consider the inverted residual blocks, \emph{i.e.}, the basic component in MobileNetV2. To do so, we replace all basic blocks (two consecutive convolutions) of ResNet26 with the inverted residual blocks as shown in Fig.~\ref{fig:bb}~and~\ref{fig:irb}. We refer to this new network as Inv-ResNet26. As shown in Fig.~\ref{fig:resnet26} and \ref{fig:invresnet}, the optimal bitwidth shifts from 1 bit to 4 bit once the basic blocks are replaced with inverted residual blocks. Thus, we can infer that the inverted residual block itself or its components are responsible for such a reversed trend.

Since an inverted residual block is composed of a point-wise convolution and a depth-wise separable convolution, we further consider the case of depth-wise separable convolution (DWSConv). To identify whether DWSConv can cause the inverted trend, we use VGG11 as a starting point and gradually replace each of the convolutions with DWSConv. We note that doing so results in architectures that gradually resemble MobileNetV1~\cite{howard2017mobilenets}. Specifically, we introduce three variants of VGG11 that have an increasing number of convolutions replaced by DWSConvs. Starting with the second layer, \emph{variant~A} has one layer replaced by DWSConv, \emph{variant~B} has four layers replaced by DWSConvs, and \emph{variant~C} has all of the layers except for the first layer replaced by DWSConvs \rudy{(the architectures are detailed in Appendix~\ref{app:arch})}.

As shown in Fig.~\ref{fig:vgg-variants}, as the number of DWSConv increases (from variant~A to variant~C), the optimal bitwidth shifts from 1~bit to 4~bits, which implies that depth-wise separable convolutions or the layers within it are affecting the optimal bitwidth. To identify which of the layers of the DWSConv (\emph{i.e.}, the depth-wise convolution or the point-wise convolution) has more impact on the optimal bitwidth, we keep the bitwidth of depth-wise convolutions fixed at 4~bits and quantize other layers. As shown in Fig.~\ref{fig:custom-quant}, the optimal curve shifts from 4~bits being the best back to 1~bit, with a similarly performing 2~bits. Thus, depth-wise convolutions appear to directly affect the optimal bitwidth trends.

\smallskip
\Centerline{\textit{\textbf{Is depth-wise convolution less resilient to quantization or}} \cpar \textit{\textbf{less sensitive to channel increase?}}}
\smallskip

\begin{wrapfigure}[17]{r}{.5\linewidth}
    % \vspace{-20pt}
    \centering
    \includegraphics[width=0.9\linewidth]{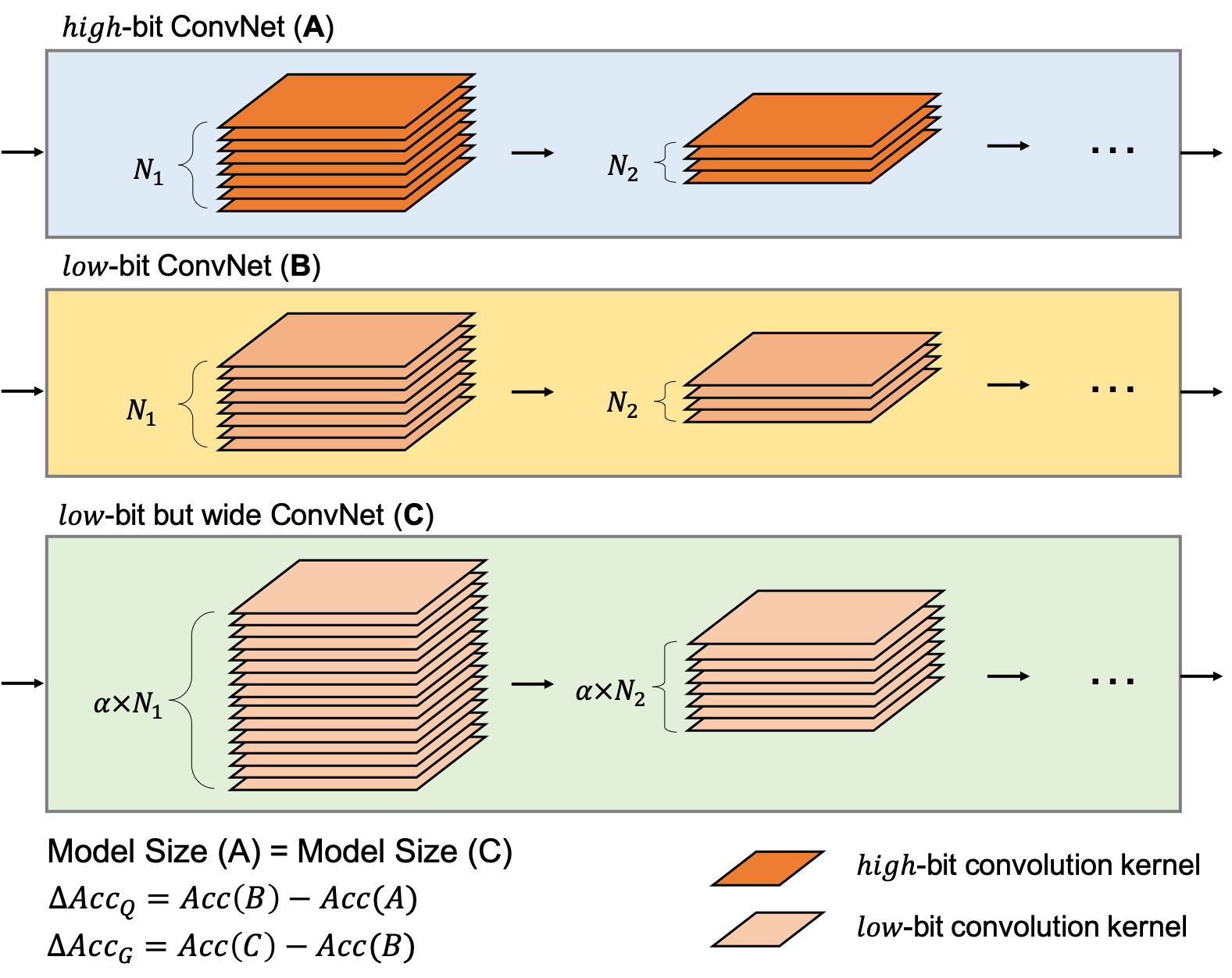}
    \caption{Visualization of our accuracy decomposition, which is used for analyzing depth-wise convolutions.}
    \label{fig:decomp}
\end{wrapfigure}

After identifying that depth-wise convolutions have a different characteristic in optimal bitwidth compared to standard all-to-all convolutions, we are interested in understanding the reason behind this. In our setup, the process to obtain a lower bitwidth network that has the same model size as a higher bitwidth network can be broken down into two steps: (1) quantize a network to lower bitwidth and (2) grow the network with width-multiplier to compensate for the reduced model size. As a result, the fact that depth-wise convolution has higher weight bitwidth better than lower weight bitwidth might potentially be due to the large accuracy degradation introduced by quantization or the small accuracy improvements from the use of more channels.

To further diagnose the cause, we decompose the accuracy difference between a lower bitwidth but wider network and a higher bitwidth but narrower network into accuracy differences incurred in the aforementioned two steps as shown in Fig.~\ref{fig:decomp}. Specifically, let $\Delta Acc_Q$ denote the accuracy difference incurred by quantizing a network and let $\Delta Acc_G$ denote the accuracy difference incurred by increasing the channel count of the quantized network.

\begin{figure*}[t]
    \centering
    \begin{subfigure}[t]{0.22\textwidth}
        \centering
        \includegraphics[height=0.8in]{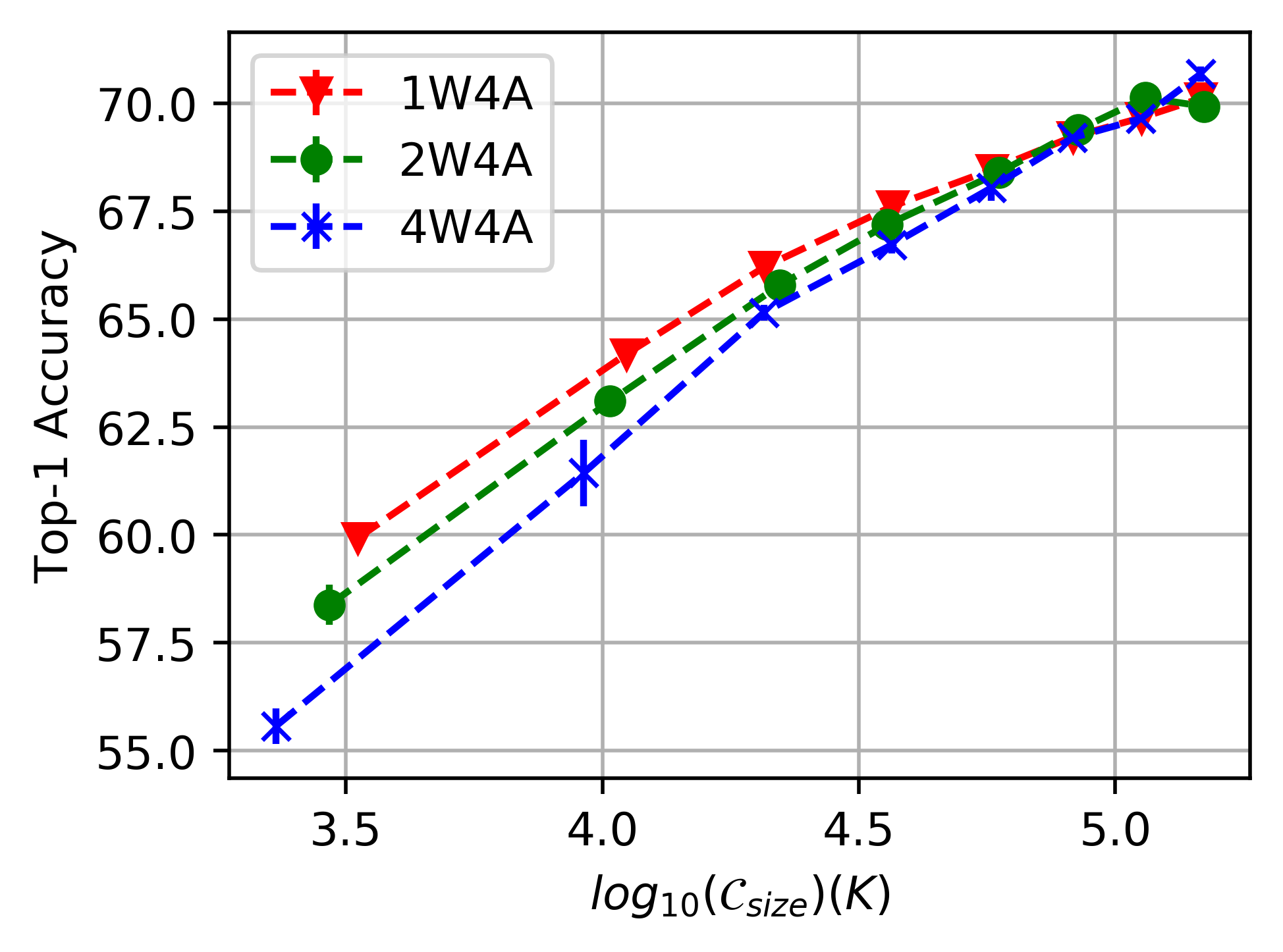}
        \caption{Variant A}
    \end{subfigure}%
    ~ 
    \begin{subfigure}[t]{0.22\textwidth}
        \centering
        \includegraphics[height=0.8in]{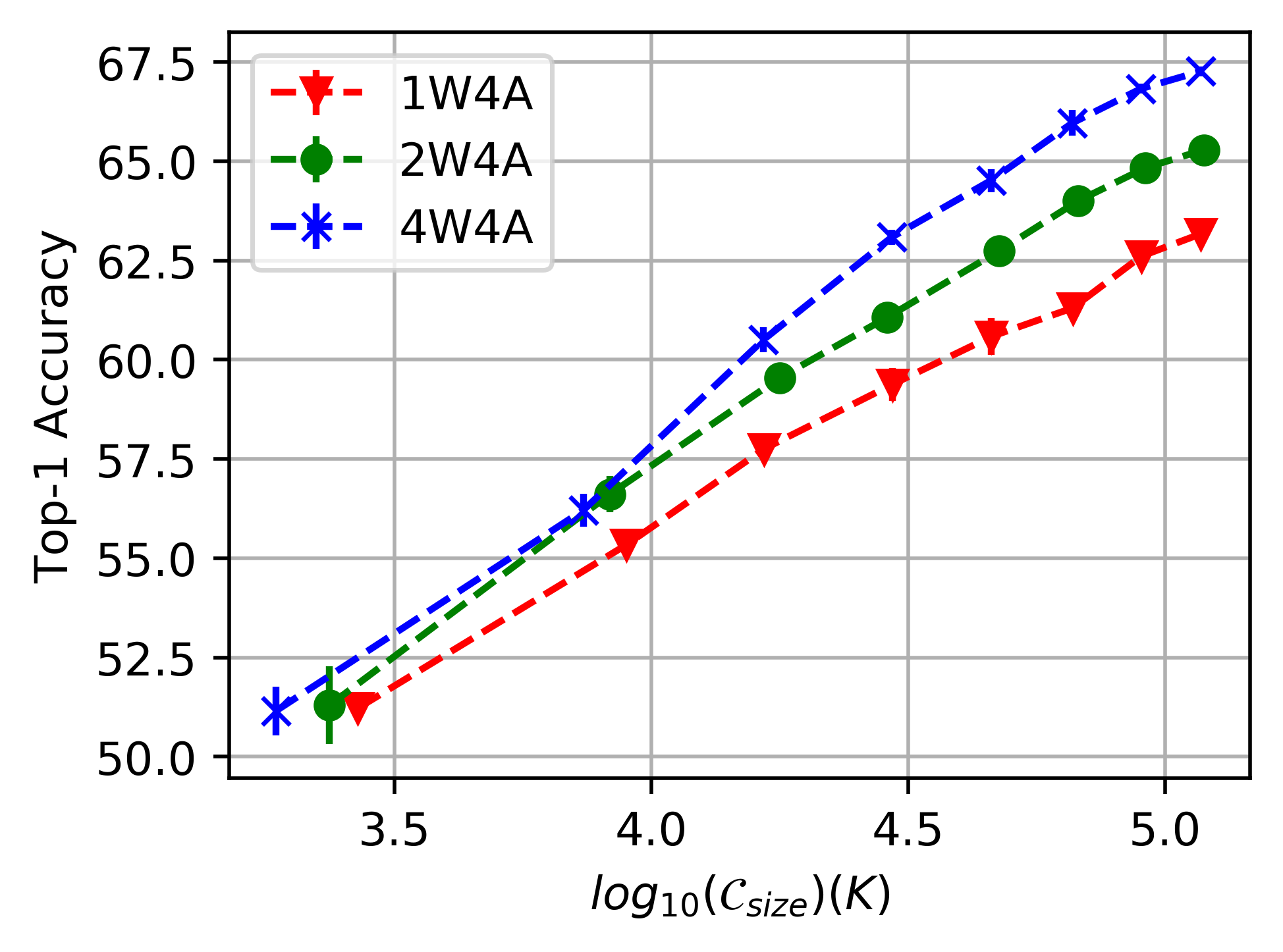}
        \caption{Variant B}
    \end{subfigure}
    ~ 
    \begin{subfigure}[t]{0.22\textwidth}
        \centering
        \includegraphics[height=0.8in]{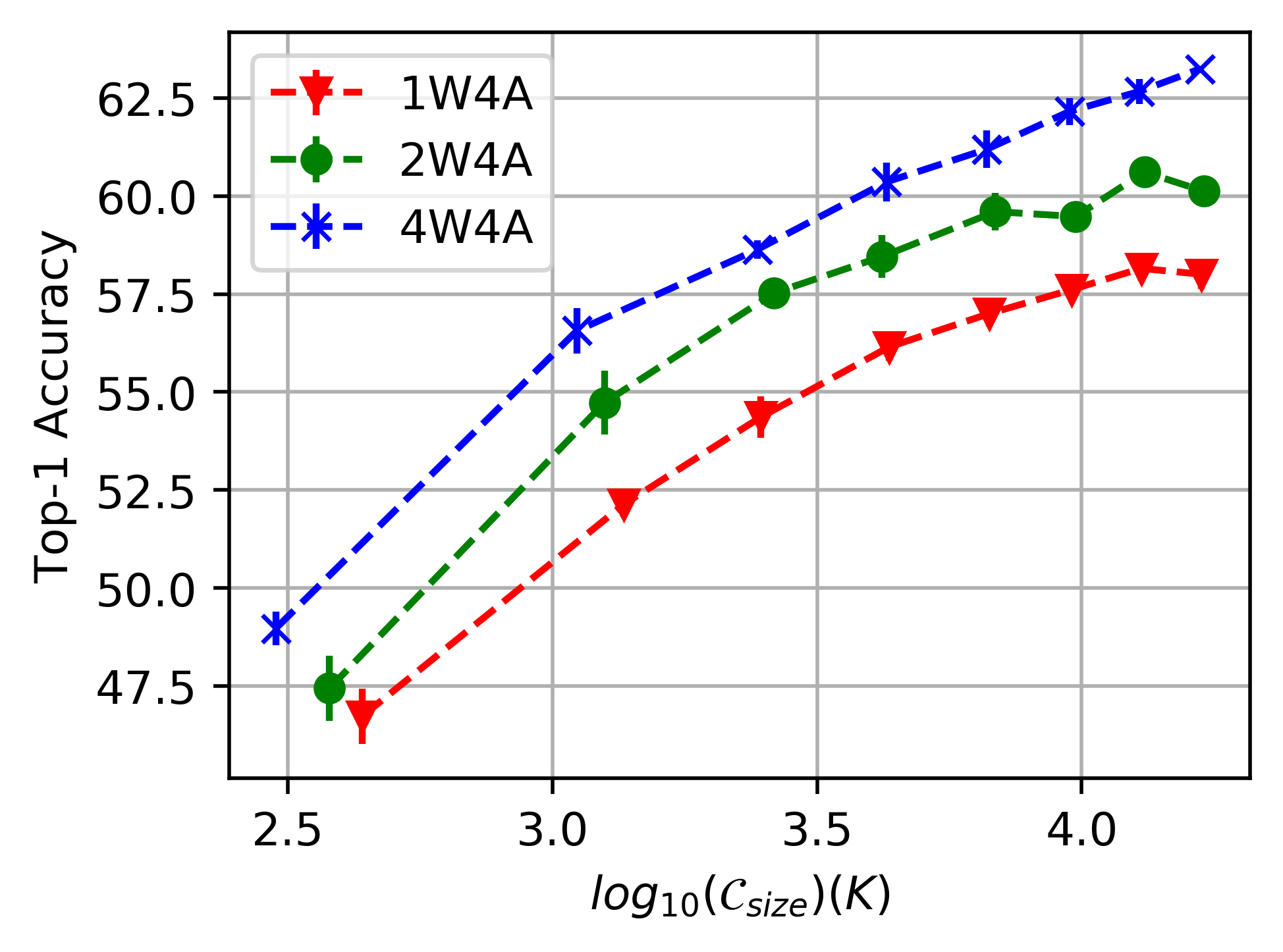}
        \caption{Variant C}
        \label{fig:variantc}
    \end{subfigure}
    ~
    \begin{subfigure}[t]{0.22\textwidth}
        \centering
        \includegraphics[height=0.8in]{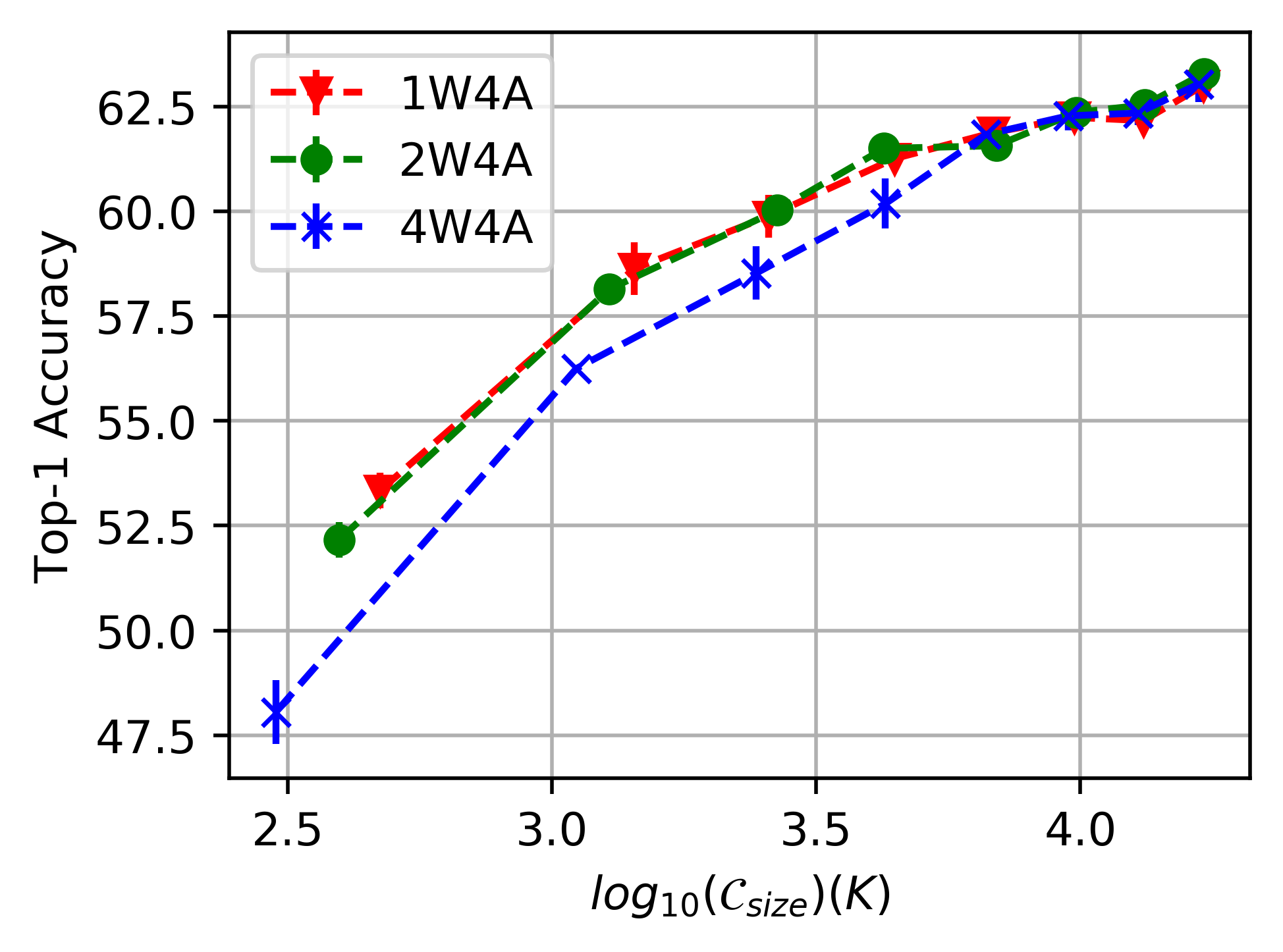}
        \caption{Variant C without quantizing depth-wise convolutions}
        \label{fig:custom-quant}
    \end{subfigure}%
    \caption{The optimal bitwidth for VGG shifts from 1 bit to 4 bit as more convolutions are replaced with depth-wise separable convolutions (DWSConv), \textit{i.e.}, from (a) to (c). Variant A, B, and C have 30\%, 60\%, and 90\% of the convolution layers replaced with DWSConv, respectively. As shown in (d), the optimal bitwidth changes back to 1 bit if we only quantize point-wise convolution but not depth-wise convolutions.}
    \label{fig:vgg-variants}
\end{figure*}

We analyze $\Delta Acc_G$ and $\Delta Acc_Q$ for networks with and without quantizing depth-wise convolutions, \emph{i.e.}, Fig.~\ref{fig:variantc} and Fig.~\ref{fig:custom-quant}. In other words, we would like to understand how depth-wise convolutions affect $\Delta Acc_G$ and $\Delta Acc_Q$. On one hand, $\Delta Acc_{Q}$ is evaluated by comparing the accuracy of the 4-bit model and the corresponding 1-bit model. On the other hand, $\Delta Acc_G$ is measured by comparing the accuracy of the 1-bit model and its 2$\times$ grown counterpart. As shown in Table~\ref{table:factor_of_pareto}, when quantizing depth-wise convolutions, $\Delta Acc_{Q}$ becomes more negative such that $\Delta Acc_{Q}~+~\Delta Acc_G~<~0$. This implies that the main reason for the optimal bitwidth change is that quantizing depth-wise convolutions introduce more accuracy degradation than it can be recovered by increasing the channel count when going below 4~bits compared to all-to-all convolutions. We note that it is expected that quantizing the depth-wise convolutions would incur smaller $\Delta Acc_{Q}$ compared to their no-quantization baseline because we essentially quantized more layers. However, depth-wise convolutions only account for 2\% of the model size but incur on average near $4\times$ more accuracy degradation when quantized.

We would like to point out that Sheng \textit{et al.}~\cite{sheng2018quantization} also find that quantizing depth-wise separable convolutions incurs large accuracy degradation. However, their results are based on post-training layer-wise quantization. As mentioned in their work~\cite{sheng2018quantization}, the quantization challenges in their setting could be resolved by quantization-aware training, which is the scheme considered in this work. Hence, our observation is novel and interesting.

\begin{table*}[t]
\caption{Quantizing depth-wise convolution introduces large accuracy degradation across model sizes. $\Delta Acc_{Q}=Acc_{1bit}-Acc_{4bit}$ denotes the accuracy introduced by quantization and $\Delta Acc_G=Acc_{1bit,2\times}-Acc_{1bit}$ denotes the accuracy improvement by increasing channel counts. The ConvNet is VGG variant~C with and without quantizing the depth-wise convolutions from 4~bits to 1~bit.}
\vskip 0.15in
\begin{center}
\begin{small}
\begin{sc}
\begin{adjustbox}{max width=1\textwidth}
\begin{tabular}{c|cc|cc|cc|cc|cc|cc}
\toprule
Width-multiplier &\multicolumn{2}{c|}{$1.00\times$}&\multicolumn{2}{c|}{$1.25\times$}&\multicolumn{2}{c|}{$1.50\times$}&\multicolumn{2}{c|}{$1.75\times$}&\multicolumn{2}{c|}{$2.00\times$}&\multicolumn{2}{c}{Average}\\
Variant~C & $\Delta Acc_{Q}$ & $\Delta Acc_G$ & $\Delta Acc_{Q}$ & $\Delta Acc_G$ & $\Delta Acc_{Q}$ & $\Delta Acc_G$ & $\Delta Acc_{Q}$ & $\Delta Acc_G$ & $\Delta Acc_{Q}$ & $\Delta Acc_G$ & $\Delta Acc_{Q}$ & $\Delta Acc_G$\\
\midrule
w/o Quantizing DWConv & -1.54 & +2.61 & -2.76 & +2.80 & -1.77 & +1.74 & -1.82 & +1.64 & -1.58 & +1.55 & -1.89 & +2.07\\
Quantizing DWConv & -8.60 & +4.39 & -7.60 & +3.41 & -7.74 & +3.19 & -8.61 & +4.09 & -7.49 & +2.25 & -8.01 & +3.47\\
\bottomrule
\end{tabular}\label{table:factor_of_pareto}
\end{adjustbox}
\end{sc}
\end{small}
\end{center}
\vskip -0.1in
\end{table*}

\smallskip
\Centerline{\textit{\textbf{Why is depth-wise convolution less resilient to quantization?}}}
\smallskip

Having uncovered that depth-wise convolutions introduce large accuracy degradation when weights are quantized below 4~bits, in this section, we investigate depth-wise convolutions from a quantization perspective. When comparing depth-wise convolutions and all-to-all convolutions in the context of quantization, they differ in the number of elements to be quantized, \emph{i.e.}, $C_{in}=1$ for depth-wise convolutions and $C_{in} >> 1$ for all-to-all convolutions.

Why does the number of elements matter? In quantization-aware training, one needs to estimate some statistics of the vector to be quantized (\emph{i.e.}, $\va$ in Equation~\ref{eq:quant-4bit} and $\bar{|\vw|}$ in Equations~\ref{eq:quant-2bit},\ref{eq:quant-1bit}) based on the elements in the vector. The number of elements affect the robustness of the estimate that further decides the quantized weights. More formally, we provide the following proposition.

\begin{proposition}\label{prop}
Let $\vw\in\R^d$ be the weight vector to be quantized where $\vw_i$ is characterized by normal distribution $\mathcal{N}(0, \sigma^2)~\forall~i$ without assuming samples are drawn independently and $d=C_{in }K_w K_h$. If the average correlation of the weights is denoted by $\rho$, the variance of $\bar{|\vw|}$ can be written as follows:

\begin{equation}\label{eq:prop}
    \begin{split}
        \Var(\bar{|\vw|}) = \frac{\sigma^2}{d} + \frac{(d-1) \rho \sigma^2}{d} - \frac{2\sigma^2}{\pi}.
    \end{split}
\end{equation}
\end{proposition}

The proof is in Appendix~\ref{app:proof}. This proposition states that, as the number of elements ($d$) increases, the variance of the estimate can be reduced (due to the first term in equation~(\ref{eq:prop})). The second term depends on the correlation between weights. Since the weights might not be independent during training, the variance is also affected by their correlations. 

\begin{wrapfigure}[12]{r}{.5\linewidth}
    \vspace{-20pt}
    \centering
    \includegraphics[width=0.8\linewidth]{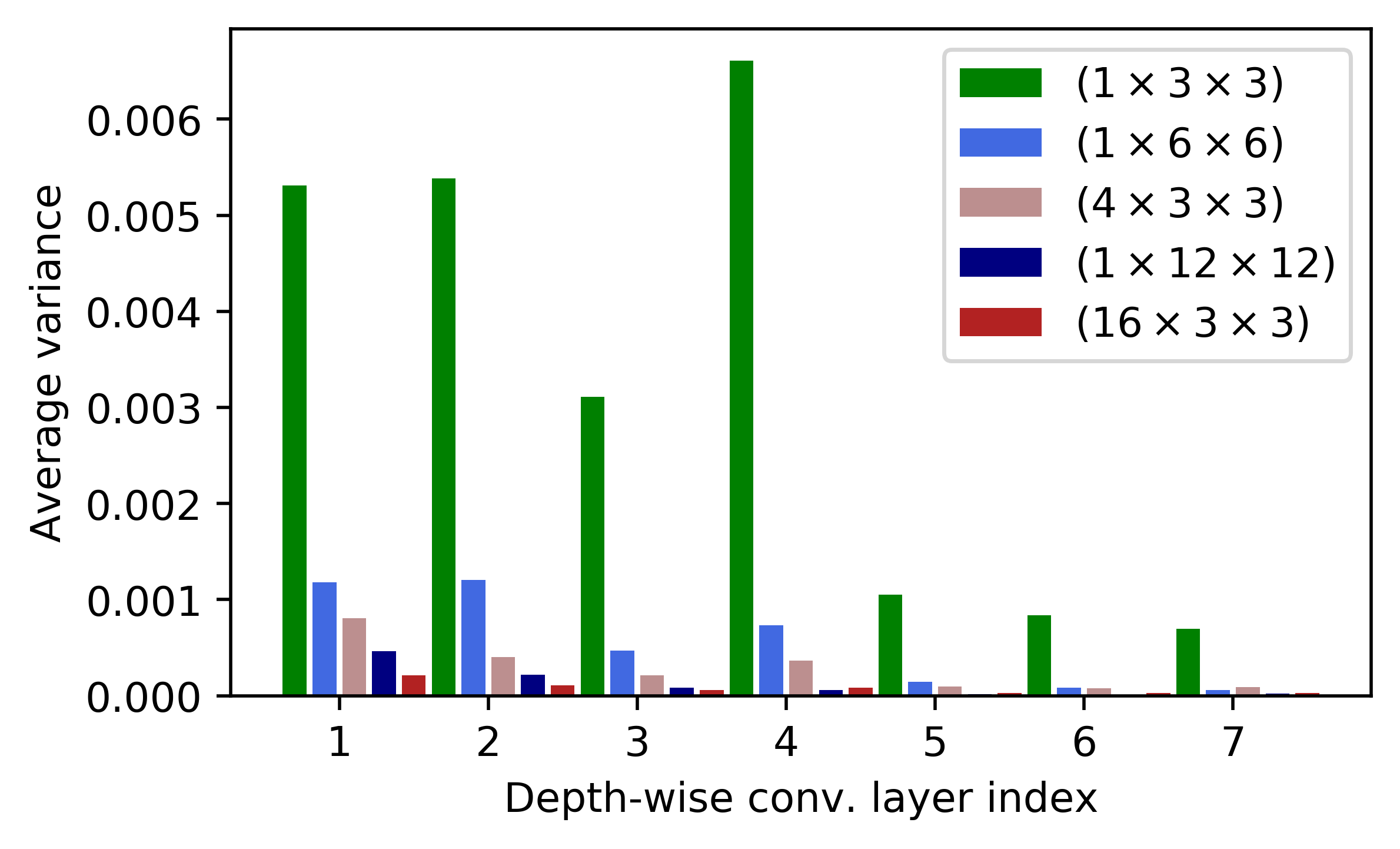}
    \caption{The average estimate $\Var(\bar{|\vw|})$ for each depth-wise convolution under different $d=(C_{in}\times K_w\times K_h$) values.}
    \label{fig:variance}
\end{wrapfigure}

We empirically validate Proposition~\ref{prop} by looking into the sample variance of $\bar{|\vw|}$ across the course of training\footnote{We treat the calculated $\bar{|\vw|}$ at each training step as a sample and calculate the sample variance across training steps.} for different $d$ values by increasing $(K_w,K_h)$ or $C_{in}$. Specifically, we consider the $0.5\times$ VGG variant~C and change the number of elements of the depth-wise convolutions. Let $d=(C_{in}\times K_w\times K_h)$ for a convolutional layer, we consider the original depth-wise convolution, \emph{i.e.}, $d=1\times 3 \times 3$ and increased channels with $d=4\times 3 \times 3$ and $d=16\times 3\times 3$, and increased kernel size with $d=1\times 6 \times 6$ and $d=1\times 12\times 12$. The numbers are selected such that increasing the channel count results in the same $d$ compared to increasing the kernel sizes. We note that when the channel count ($C_{in}$) is increased, it is no longer a depth-wise convolution, but rather a group convolution.

In Fig.~\ref{fig:variance}, we analyze layer-level sample variance by averaging the kernel-level sample variance in the same layer. First, we observe that results align with Proposition~\ref{prop}. That is, one can reduce the variance of the estimate by increasing the number of elements along both the channel ($C_{in}$) and kernel size dimensions ($K_{w},K_{h}$). Second, we find that increasing the number of channels ($C_{in}$) is more effective in reducing the variance than increasing kernel size ($K_{w},K_{h}$), which could be due to the weight correlation, \emph{i.e.}, intra-channel weights have larger correlation than inter-channel weights.

\begin{wrapfigure}[17]{l}{.45\linewidth}
    % \vspace{-20pt}
    \centering
    \includegraphics[width=0.9\linewidth]{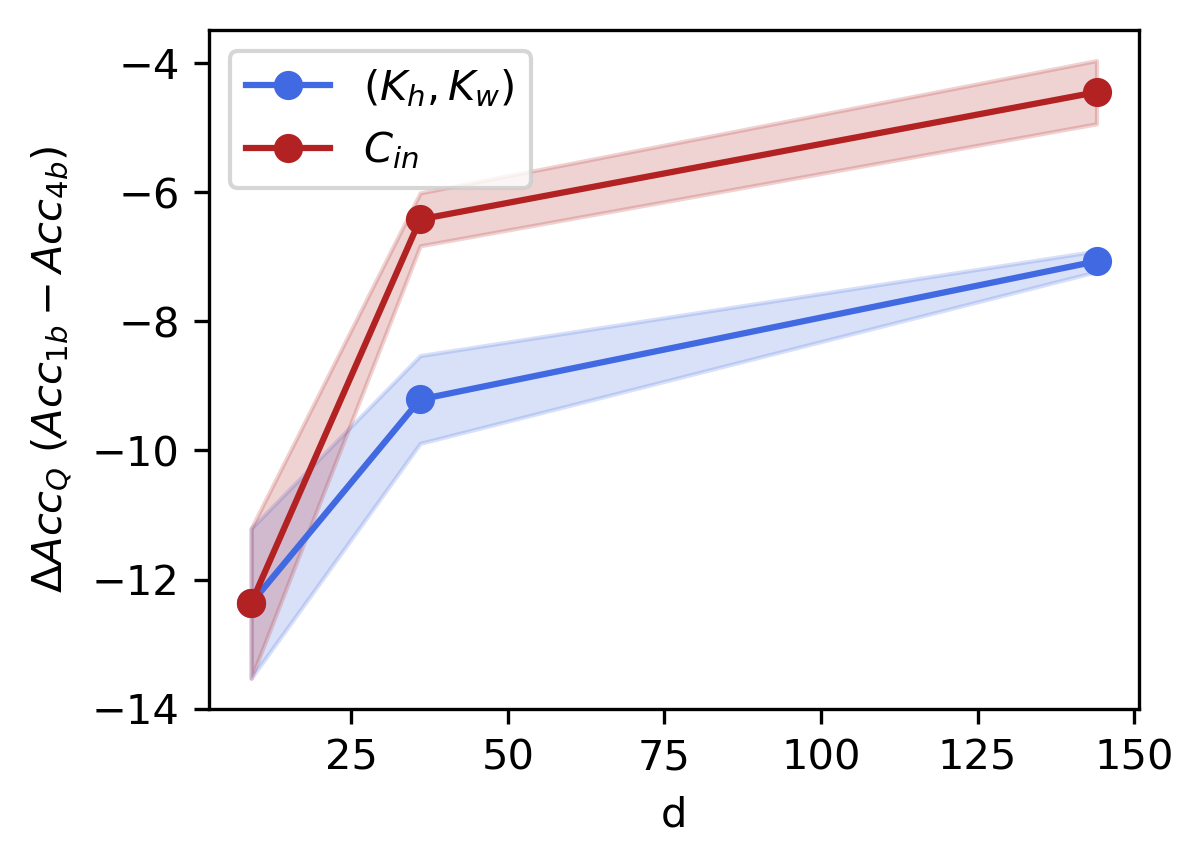}
    \caption{\rudy{$d$ negatively correlates with the variance and positively correlates with the accuracy difference induced by quantization $\Delta Acc_{Q}=Acc_{1bit}-Acc_{4bit}$.}}
    \label{fig:acc-q}
\end{wrapfigure}

Nonetheless, while lower variance suggests a more stable value during training, it might not necessarily imply lower quantization error for the quantized models. Thus, we conduct an accuracy sensitivity analysis with respect to quantization for different $d$ values. More specifically, we want to understand how $d$ affects the accuracy difference between lower bitwidth (1~bit) and higher bitwidth (4~bits) models ($\Delta Acc_{Q}$). As shown in Fig.~\ref{fig:acc-q}, we empirically find that $d$ positively correlates with $\Delta Acc_{Q}$, \textit{i.e.}, the larger the $d$, the smaller the accuracy degradation is. On the other hand, when comparing channel counts and kernel sizes, we observe that increasing the number of channels is more effective than increasing the kernel size in reducing accuracy degradation caused by quantization. This analysis sheds light on the two different trends observed in Fig.~\ref{fig:cifar_pareto}.

\subsection{Remarks and scaling up to ImageNet}\label{sec:third_finding}
We have two intriguing findings so far. First, there exists some bitwidth that is better than others across model sizes when compared under a given model size. Second, the optimal bitwidth is architecture-dependent. More specifically, the optimal weight bitwidth negatively correlates with the fan-in channel counts per convolutional kernel. These findings show promising results for the hardware and software researchers to support only a certain set of bitwidths when it comes to parameter-efficiency. For example, \rudy{use} binary weights for networks with all-to-all convolutions.

Next, we scale up our analysis to the ImageNet dataset. Specifically, we study ResNet50 and MobileNetV2 on the ImageNet dataset. Since we keep the bitwidth of the first and last layer quantized at 8~bits, scaling them in terms of width will grow the number of parameters much more quickly than other layers. As a result, we keep the number of channels for the first and last channel fixed for the ImageNet experiments. As demonstrated in Section~\ref{sec:first_finding}, the bit ordering is consistent across model sizes, we conduct our analysis for ResNet50 and MobileNetV2 by scaling them down with a width-multiplier of $0.25\times$ for computational considerations. The choices of bitwidths are limited to $\{1,2,4,8\}$.

\begin{table*}[t!]
\caption{bitwidth ordering for MobileNetV2 and ResNet50 with \emph{the model size aligned to the $0.25\times$ 8~bits models} on ImageNet. Each cell reports the top-1 accuracy of the corresponding model. The trend for the optimal bitwidth is similar to that of CIFAR-100 (4 bit for MobileNetV2 and 1 bit for ResNet).}
\vskip 0.15in
\begin{center}
\begin{small}
\begin{sc}
\begin{adjustbox}{max width=1\textwidth}
\begin{tabular}{c|cccc|c}
\toprule

Weight bitwidth for &\multicolumn{4}{c}{MobileNetV2}&\multicolumn{1}{|c}{ResNet50}\\
Convs $\backslash$ DWConvs & 8~bits & 4~bits & 2~bits & 1~bit & None \\
\midrule
8~bits& 52.17 & 53.89 & 50.51 & 48.78 & 71.11\\
4~bits& 56.84 & \textbf{59.51} & 57.37 & 55.91 & 74.65\\
2~bits& 53.89 & 57.10 & 55.26 & 54.04 & 75.12\\
1~bit& 54.82 & 58.16 & 56.90 & 55.82 & \textbf{75.44}\\
\bottomrule
\end{tabular}\label{table:grid-search}
\end{adjustbox}
\end{sc}
\end{small}
\end{center}
\vskip -0.1in
\end{table*}
As shown in Table~\ref{table:grid-search}, we can observe a trend similar to the CIFAR-100 experiments, \emph{i.e.}, for networks without depth-wise convolutions, the lower weight bitwidths the better, and for networks with depth-wise convolutions, there are sweet spots for depth-wise and other convolutions. Specifically, the final weight bitwidth selected for MobileNetV2 is 4~bits for both depth-wise and standard convolutions. On the other hand, the selected weight bitwidth for ResNet50 is 1~bit. If bit ordering is indeed consistent across model sizes, these results suggest that the optimal bitwidth for MobileNetV2 is 4 bit and it is 1 bit for ResNet50. However, throughout our analysis, we have not considered mixed-precision, which makes it unclear if the so-called optimal bitwidth (4 bit for MobileNetV2 and 1 bit for ResNet-50) is still optimal when compared to mixed-precision quantization.

As a result, we further compare with mixed-precision quantization that uses reinforcement learning to find the layer-wise bitwidth~\cite{wang2019haq}. Specifically, we follow~\cite{wang2019haq} and use a reinforcement learning approach to search for the lowest bitwidths without accuracy degradation (compared to the 8~bits fixed point models). To compare the searched model with other alternatives, we use width-multipliers on top of the searched network match the model size of the 8 bit quantized model. We consider networks of three sizes, \emph{i.e.}, the size of $1\times, 0.5\times$ and $0.25\times$ 8-bit fixed point models. As shown in Table~\ref{table:imagenet}, we find that a single bitwidth (selected via Table~\ref{table:grid-search}) outperforms both 8 bit quantization and mixed-precision quantization by a significant margin for both networks considered. This results suggest that searching for the bitwidth without accuracy degradation is indeed a sub-optimal strategy and can be improved by incorporating channel counts into the search space and reformulate the optimization problem as maximizing accuracy under storage constraints. Moreover, our results also imply that when the number of channels are allowed to be altered, a single weight bitwidth throughout the network shows great potential for model compression, which has the potential of greatly reducing the software and hardware optimization costs for quantized ConvNets.  

\begin{table*}[t!]
\caption{The optimal bitwidth selected in Table~\ref{table:grid-search} is indeed better than 8 bit when scaled to larger model sizes and more surprisingly, it is better than mixed-precision quantization. All the activations are quantized to 8~bits.}
\vskip 0.15in
\begin{center}
\begin{small}
\begin{sc}
\begin{adjustbox}{max width=1\textwidth}
\begin{tabular}{cc|cc|cc|cc}
\toprule
\multicolumn{2}{c|}{Width-multiplier for 8-bit model} & \multicolumn{2}{c|}{$1\times$} & \multicolumn{2}{c|}{$0.5\times$} & \multicolumn{2}{c}{$0.25\times$}\\
\midrule
Networks & Methods & Top-1 (\%) & $\mathcal{C}_{size}~(10^6)$ & Top-1 (\%) & $\mathcal{C}_{size}~(10^6)$ & Top-1 (\%) & $\mathcal{C}_{size}~(10^6)$\\
\midrule
\multirow{4}{*}{ResNet50}&Floating-point & 76.71 & 816.72 & 74.71 & 411.48 & 71.27 & 255.4\\
&8~bits & 76.70 & 204.18 & 74.86 & 102.87 & 71.11 & 63.85\\
&Flexible~\cite{wang2019haq} & 77.23 & 204.18 & 76.04 & 102.90 & 74.30 & 63.60\\
&\textit{Optimal} (1 bit) & \textbf{77.58} & 204.08 & \textbf{76.70} & 102.83 & \textbf{75.44} & 63.13\\
\midrule
\multirow{4}{*}{MobileNetV2}&Floating-point & 71.78 & 110.00 & 63.96 & 61.76 & 52.79 & 47.96\\
&8~bits & 71.73 & 27.50 & 64.39 & 15.44 & 52.17 & 11.99\\
&Flexible~\cite{wang2019haq} & 72.13 & 27.71 & 65.00 & 15.54 & 55.20 & 12.10\\
&\textit{Optimal} (4 bit) & \textbf{73.91} & 27.56 & \textbf{68.01} & 15.53 & \textbf{59.51} & 12.15\\
\bottomrule
\end{tabular}\label{table:imagenet}
\end{adjustbox}
\end{sc}
\end{small}
\end{center}
\vskip -0.1in
\end{table*}

\section{Conclusion}\label{sec:conclusion}
In this work, we provide the first attempt to understand the ordering between different weight bitwidths by allowing the channel counts of the considered networks to vary using the width-multiplier. If there exists such an ordering, it may be helpful to focus on software/hardware support for higher-ranked bitwidth when it comes to parameter-efficiency, which in turn reduces software/hardware optimization costs. To this end, we have three surprising findings: (1) there exists a weight bitwidth that is better than others across model sizes under a given model size constraint, (2) the optimal weight bitwidth of a convolutional layer negatively correlates to the fan-in channel counts per convolutional kernel, and (3) with a single weight bitwidth for the whole network, one can find configurations that outperform layer-wise mixed-precision quantization using reinforcement learning when compared under a given same model size constraint. Our results suggest that when the number of channels are allowed to be altered, a single weight bitwidth throughout the network shows great potential for model compression.

\section*{Acknowledgement}
This research was supported in part by NSF CCF Grant No. 1815899, NSF CSR Grant No. 1815780, and NSF ACI Grant No. 1445606 at the Pittsburgh Supercomputing Center (PSC).

\bibliographystyle{splncs04}
\bibliography{example_paper}

\newpage
\appendix
\section{Clipping Point for Quantization-aware Training}\label{app:alpha}
\begin{figure*}[h]
    \centering
    \includegraphics[width=0.8\textwidth]{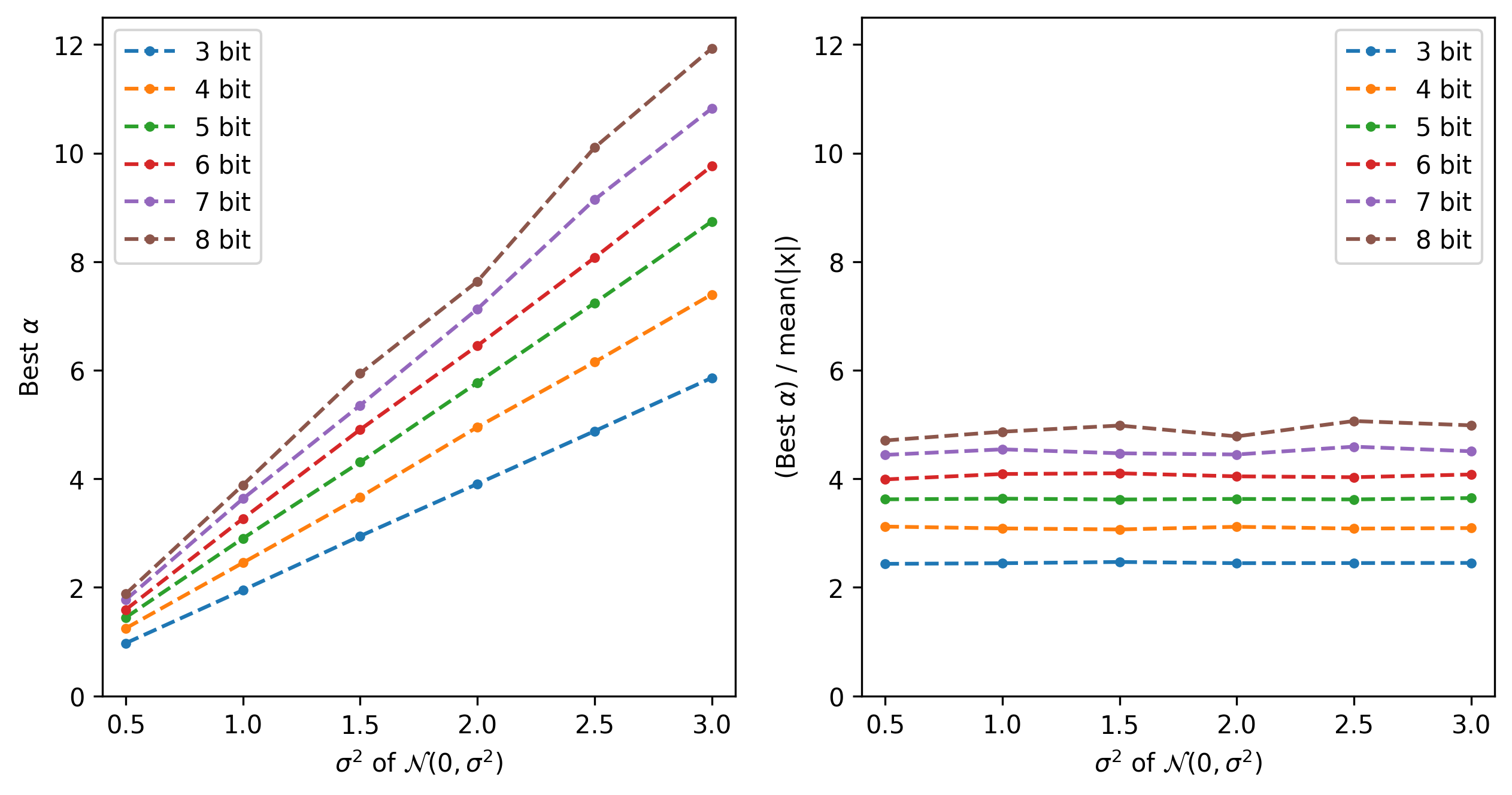}
    \caption{Finding best $\eva_i$ for different precision values empirically through simulation using Gaussian with various $\sigma^2$.}
    \label{fig:alpha-search}
\end{figure*}

As mentioned earlier, $\va \in \R^{C_{out}}$ denotes the vector of clipping factors which is selected to minimize $\lVert Q(\mW_{i,:}) - \mW_{i,:} \rVert^2_2$ by assuming $\mW_{i,:}~\sim \mathcal{N}(0, \sigma^2\mI)$. More specifically, we run simulations for weights drawn from a zero-mean Gaussian distribution with several variances and identify the best $\eva_i^*=\arg \min_{\eva_i}\lVert Q_{\eva_i}(\mW_{i,:}) - \mW_{i,:} \rVert^2_2$ empirically. According to our simulation, we find that one can infer $\eva_i$ from the sample mean $\bar{|\mW_{i,:}|}$, which is shown in Fig.~\ref{fig:alpha-search}. As a result, for the different precision values considered, we find  $c = \frac{\bar{|\mW_{i,:}|}}{\eva_i^*}$ via simulation and use the obtained $c$ to calculate $\eva_i$ on-the-fly throughout training.

\section{Network Architectures}\label{app:arch}
\rudy{For the experiments in Section~\ref{sec:first_finding}, the ResNets used are detailed in Table~\ref{table:resnet_arch}. Specifically, for the points in Fig.~\ref{fig:cifar_resnet}, we consider ResNet20 to ResNet56 with width-multipliers of $0.5\times, 1\times, 1.5\times$, and $2\times$ for the 4-bit case. Based on these values, we consider additional width-multipliers $2.4\times$ and $2.8\times$ for the 2-bit case and $2.5\times, 3\times, 3.5\times,$ and $3.9\times$ for the 1-bit case. We note that the right-most points in Fig.~\ref{fig:cifar_resnet} is a $10\times$ ResNet26 for the 4~bits case. On the other hand, VGG11 is detailed in Table~\ref{table:vgg_arch} for which we consider width-multipliers from $0.25\times$ to $2\times$ with a step of $0.25$ for the 4~bits case (blue dots in Fig.~\ref{fig:cifar_vggnet}). The architecture of MobileNetV2 used in the CIFAR-100 experiments follows the original MobileNetV2~(Table 2 in \cite{sandler2018mobilenetv2}) but we change the stride of all the bottleneck blocks to 1 except for the fifth bottleneck block, which has a stride of 2. As a result, we down-sample the image twice in total, which resembles the ResNet design for the CIFAR experiments~\cite{he2016deep}. Similar to VGG11, we consider width-multipliers from $0.25\times$ to $2\times$ with a step of $0.25$ for MobileNetV2 for the 4~bits case (blue dots in Fig.~\ref{fig:cifar_mbnet}).}

\begin{table*}[h]
\caption{ResNet20 to ResNet56}
\vskip 0.15in
\begin{center}
\begin{small}
\begin{sc}
\begin{adjustbox}{max width=1\textwidth}
\begin{tabular}{c|c|c|c|c|c|c|c}
\toprule
Layers & 20 & 26 & 32 & 38 & 44 & 50 & 56\\
\midrule
Stem & \multicolumn{7}{c}{$\text{Conv2d (16,3,3)} \text{ Stride } 1$}\\
\midrule
Stage 1 & $3\times  \begin{cases}
      \text{Conv2d} (16,3,3) \text{ Stride } 1\\
      \text{Conv2d} (16,3,3) \text{ Stride } 1
    \end{cases}$ & $4\times$ & $5\times$ & $6\times$ & $7\times$ & $8\times$ & $9\times$\\
\midrule
Stage 2 & $3\times  \begin{cases}
      \text{Conv2d} (32,3,3) \text{ Stride } 2\\
      \text{Conv2d} (32,3,3) \text{ Stride } 1
    \end{cases}$ & $4\times$ & $5\times$ & $6\times$ & $7\times$ & $8\times$ & $9\times$\\
\midrule
Stage 3 & $3\times  \begin{cases}
      \text{Conv2d} (64,3,3) \text{ Stride } 2\\
      \text{Conv2d} (64,3,3) \text{ Stride } 1
    \end{cases}$ & $4\times$ & $5\times$ & $6\times$ & $7\times$ & $8\times$ & $9\times$\\
\bottomrule
\end{tabular}\label{table:resnet_arch}
\end{adjustbox}
\end{sc}
\end{small}
\end{center}
\vskip -0.1in
\end{table*}

\begin{table*}[h]
\caption{Inv-ResNet26}
\vskip 0.15in
\begin{center}
\begin{small}
\begin{sc}
\begin{adjustbox}{max width=1\textwidth}
\begin{tabular}{c|c}
\toprule
Stem & $\text{Conv2d (16,3,3)} \text{ Stride } 1$\\
\midrule
Stage 1 & $4\times  \begin{cases}
      \text{Conv2d} (16\times6,1,1) \text{ Stride } 1\\
      \text{DWConv2d} (16\times6,3,3) \text{ Stride } 1\\
      \text{Conv2d} (16,1,1) \text{ Stride } 1
    \end{cases}$\\
\midrule
Stage 2 & $4\times  \begin{cases}
      \text{Conv2d} (32\times6,1,1) \text{ Stride } 1\\
      \text{DWConv2d} (32\times6,3,3) \text{ Stride } 2\\
      \text{Conv2d} (32,1,1) \text{ Stride } 1
    \end{cases}$\\
\midrule
Stage 3 & $4\times  \begin{cases}
      \text{Conv2d} (64\times6,1,1) \text{ Stride } 1\\
      \text{DWConv2d} (64\times6,3,3) \text{ Stride } 2\\
      \text{Conv2d} (64,1,1) \text{ Stride } 1
    \end{cases}$\\
\bottomrule
\end{tabular}\label{table:invresnet_arch}
\end{adjustbox}
\end{sc}
\end{small}
\end{center}
\vskip -0.1in
\end{table*}

\begin{table*}[h]
\caption{VGGs}
\vskip 0.15in
\begin{center}
\begin{small}
\begin{sc}
\begin{adjustbox}{max width=1\textwidth}
\begin{tabular}{c|c|c|c}
\toprule
VGG11 & Variant A & Variant B & Variant C\\
\midrule
\multicolumn{4}{c}{$\text{Conv2d (64,3,3)}$}\\
\midrule
\multicolumn{4}{c}{MaxPooling}\\
\midrule
$\text{Conv2d (128,3,3)}$ & $\begin{cases}
      \text{Conv2d} (128,1,1)\\
      \text{DWConv2d} (128,3,3)
    \end{cases}$ & $\begin{cases}
      \text{Conv2d} (128,1,1)\\
      \text{DWConv2d} (128,3,3)
    \end{cases}$ & $\begin{cases}
      \text{Conv2d} (128,1,1)\\
      \text{DWConv2d} (128,3,3)
    \end{cases}$\\
\midrule
\multicolumn{4}{c}{MaxPooling}\\
\midrule
$\text{Conv2d (256,3,3)}$&$\text{Conv2d (256,3,3)}$&$\begin{cases}
      \text{Conv2d} (256,1,1)\\
      \text{DWConv2d} (256,3,3)
    \end{cases}$&$\begin{cases}
      \text{Conv2d} (256,1,1)\\
      \text{DWConv2d} (256,3,3)
    \end{cases}$\\
\midrule
$\text{Conv2d (256,3,3)}$&$\text{Conv2d (256,3,3)}$&$\begin{cases}
      \text{Conv2d} (256,1,1)\\
      \text{DWConv2d} (256,3,3)
    \end{cases}$&$\begin{cases}
      \text{Conv2d} (256,1,1)\\
      \text{DWConv2d} (256,3,3)
    \end{cases}$\\
\midrule
\multicolumn{4}{c}{MaxPooling}\\
\midrule
$\text{Conv2d (512,3,3)}$&$\text{Conv2d (512,3,3)}$&$\begin{cases}
      \text{Conv2d} (512,1,1)\\
      \text{DWConv2d} (512,3,3)
    \end{cases}$&$\begin{cases}
      \text{Conv2d} (512,1,1)\\
      \text{DWConv2d} (512,3,3)
    \end{cases}$\\
\midrule
$\text{Conv2d (512,3,3)}$&$\text{Conv2d (512,3,3)}$&$\text{Conv2d (512,3,3)}$&$\begin{cases}
      \text{Conv2d} (512,1,1)\\
      \text{DWConv2d} (512,3,3)
    \end{cases}$\\
\midrule
\multicolumn{4}{c}{MaxPooling}\\
\midrule
$\text{Conv2d (512,3,3)}$&$\text{Conv2d (512,3,3)}$&$\text{Conv2d (512,3,3)}$&$\begin{cases}
      \text{Conv2d} (512,1,1)\\
      \text{DWConv2d} (512,3,3)
    \end{cases}$\\
\midrule
$\text{Conv2d (512,3,3)}$&$\text{Conv2d (512,3,3)}$&$\text{Conv2d (512,3,3)}$&$\begin{cases}
      \text{Conv2d} (512,1,1)\\
      \text{DWConv2d} (512,3,3)
    \end{cases}$\\
\midrule
\multicolumn{4}{c}{MaxPooling}\\
\bottomrule
\end{tabular}\label{table:vgg_arch}
\end{adjustbox}
\end{sc}
\end{small}
\end{center}
\vskip -0.1in
\end{table*}

\section{Proof for Proposition 5.1}\label{app:proof}
Based on the definition of variance, we have:
\begin{align*}
    \Var(\frac{1}{d}\sum_{i=1}^d |\vw_i|) &:= \E \left[\left( \frac{1}{d}\sum_{i=1}^d |\vw_i| \right)^2 - \left( \E \frac{1}{d}\sum_{i=1}^d |\vw_i| \right)^2\right]\\
    &= \E \left[\left( \frac{1}{d}\sum_{i=1}^d |\vw_i| \right)^2 - \frac{2\sigma^2}{\pi} \right]\\
    &= \frac{1}{d^2} \E \left( \sum_{i=1}^d |\vw_i| \right)^2 - \frac{2\sigma^2}{\pi}\\
    &= \frac{\sigma^2}{d} + \frac{d-1}{d} \rho \sigma^2 - \frac{2\sigma^2}{\pi}.
\end{align*}

\end{document}

%% file: math_commands.tex
\usepackage{amsmath,amsfonts,bm}

\def\eqref#1{equation~\ref{#1}}
\def\1{\bm{1}}

\newcommand{\test}{\mathcal{D_{\mathrm{test}}}}

\def\va{{\bm{a}}}

\def\vw{{\bm{w}}}

\def\eva{{a}}

\def\evr{{r}}

\def\mI{{\bm{I}}}

\def\mM{{\bm{M}}}

\def\mW{{\bm{W}}}

\DeclareMathAlphabet{\mathsfit}{\encodingdefault}{\sfdefault}{m}{sl}
\SetMathAlphabet{\mathsfit}{bold}{\encodingdefault}{\sfdefault}{bx}{n}

\newcommand{\E}{\mathbb{E}}

\newcommand{\R}{\mathbb{R}}

\newcommand{\Var}{\mathrm{Var}}